# COMPLETE END-TO-END LOW COST SOLUTION TO A 3D SCANNING SYSTEM WITH INTEGRATED TURNTABLE

SAED KHAWALDEH *

*Erasmus+ Joint Master Program in Medical Imaging and Applications*
*University of Burgundy (France), University of Cassino (Italy) and University of Girona (Spain)*
*khawaldeh.saed@gmail.com*

TAJWAR ABRAR ALEEF *

*Erasmus+ Joint Master Program in Medical Imaging and Applications*
*University of Burgundy (France), University of Cassino (Italy) and University of Girona (Spain)*
*tajwar_aleef@etu.u-bourgogne.fr*

USAMA PERVAIZ *

*Erasmus+ Joint Master Program in Medical Imaging and Applications*
*University of Burgundy (France), University of Cassino (Italy) and University of Girona (Spain)*
*12beeupervaiz@seecs.edu.pk*

VU HOANG MINH *

*Erasmus+ Joint Master Program in Medical Imaging and Applications*
*University of Burgundy (France), University of Cassino (Italy) and University of Girona (Spain)*
*hoangminh.vu@smartdatics.com*

YEMAN BRHANE HAGOS *

*Erasmus+ Joint Master Program in Medical Imaging and Applications*
*University of Burgundy (France), University of Cassino (Italy) and University of Girona (Spain)*
*yemanbrhane1989@gmail.com*

3D reconstruction is a technique used in computer vision and it has a wide range of applications in areas like object recognition, city modelling, virtual reality, physical simulations, video games and special effects. Previously, to perform a 3D reconstruction, specialized hardware was required. Such system was often very expensive and was only available for industrial or research purpose. Nowadays, with the rise of high-quality 3D scanners available at low price, it is possible to design complete 3D scanning systems at very low cost. The objective of this work is to design a homemade acquisition and processing system to perform 3D scanning and reconstruction of objects. The goal of this work also includes making the 3D scanning process fully automated by building and integrating a turntable alongside the software. In addition, the user is able to perform a full 3D scan by the press of a few buttons on our dedicated Graphical User Interface (GUI) which has been designed for this purpose. Hence, the product of our work will be an acquisition and a processing software capable of controlling the turning table, acquire point cloud frames, register them and reconstruct the 3D mesh which can be exported afterwards to a 3D printer. To achieve this goal, three main steps were required. First, our system acquires point cloud data of a person/object using inexpensive camera sensor. Second, align and convert the acquired point cloud data into a watertight mesh of good quality. Third, export the reconstructed model to a 3D printer to obtain a proper 3D print of the model.

*Keywords: 3D Body Scanning, 3D Printing, 3D Reconstruction, Iterative Closest Process, Automated Scanning System, Kinect v2.0 Sensor, RGB-D camera, Point Cloud Library (PCL).*

\* Authors contributed equally to the work

## 1. Introduction

Three Dimensional printing technology has made a breakthrough in recent years as many applications are being developed based around it. The ability to design a digital concept and physically print the results of that concept is a promising technology that will only become more commonplace as it continues to be refined [1]. The combination of both technologies, 3D reconstruction and 3D printing, into a general 3D copy machine bears an enormous economical potential for the future [2]. In paper [3], authors proposed that for obtaining a full human body scan with good quality, three camera sensors must be used. In their work, two of these sensors were used to record the raw data for the upper and lower parts of the body while ensuring that the areas of scanning for the different sensors do not overlap in order to avoid complications in registering process. Additionally, a third is sensor fixed opposite to the two ones to obtain data for the middle part of the body that is not covered by the first two sensors. The results are convincing, but the use of three cameras is not only costly but also requires calibration and hence complicate the process of acquisition. In paper [4], a single sensor based 3D scanning system is presented. An amount of nonrigid deformation is permitted while performing the scanning process which guaranteed getting a high-quality output avoiding motion artefacts. No dependency on any pre knowledge about the shape of an object to be scanned is needed in this paper. To ensure high-quality scanning results, loop closures were detected which distributed the alignment error over the whole loop, and lastly, bundle adjustment algorithm is applied to optimize for the latent 3D shape and nonrigid deformation parameters at the same time. As a result, high-quality scanning outcome was obtained even in challenging scanning conditions. In paper [5], an analysis of error related to the depth of the sensor, effects caused by increasing or decreasing the distance to object, quality and density of point clouds obtained is presented. Moreover, the paper proposes a model (with different calibration parameters to be set) which enable acquiring the XYZ coordinates of the 3D data. Based on the results obtained, they suggest using an accurate calibration of both IR camera and RGB cameras to get rid of distortions in the point cloud and to reduce the misalignments between the colour and depth data. Their results show that the random error of depth measurements increases quadratically with increasing distance from the sensor, and the depth resolution decreases quadratically with increasing distance from the sensor. In paper [6], a human body reconstruction method depending on a noisy monocular image and range data acquired using Microsoft Kinect sensor is proposed. The 3D shape is estimated by integrating multiple monocular views of a human subject moving in front of the camera sensor. To solve the problem related to varying body pose, authors use a SCAPE body model which expresses the 3D body shape in different mathematical quantities and detect the variations occurring. Further, they explain a novel method to reduce the distance between the projected 3D body contour and the image silhouette which depends on analytic derivatives of the objective function. They suggest a simple approach to estimate body measurements from the recovered SCAPE model and then they calculate the accuracy of the method they used in their approach to prove that it is competitive compared to the commercial body scanning systems available in the market.

To this date, there are several softwares already available that can take 3D scans and export models for 3D printing using various types of sensors and various algorithms to get to the end results. Softwares such 'ReconstructMe', 'Skanect', 'Microsoft 3D Scan' uses Kinect as their primary sensor. Hence, we will be making a product similar to them. Furthermore, the target of this research work was to develop something as close to Skanect. The main motivation of this work was to develop a total low-cost system for 3D scanning. Using open source libraries allows our complete software to be free of charge to the consumers and the use of cheap sensors makes the whole system cost effective. To make the system even more easy to use, integration was done with a low-cost turntable to finally output a total low-cost end-to-end solution for 3D scanning. Compared to other available software, most require a purchase and either expensive sensors or expensive turntables to output and print scans.

In this work, an end-to-end 3D reconstruction software was developed using Kinect and Point Cloud Library (PCL) for scanning, reconstructing and exporting a 3D model of any object into a 3D printer. First, a brief introduction to 3D reconstruction and an explanation of the depth measuring technology utilized by the Kinect v2.0 is given. This project is based on open source libraries and frameworks, thus, a presentation of the tools and libraries utilized during the design of the software is provided. The main library used for this project is the Point Cloud Library (PCL). Being an open source library, free for commercial use and with wide contents of online documentation, PCL is one of the key core elements for the development of this project. Regarding the Graphical User Interface (GUI) built, some images are included in order to explain how the user can perform a new scan or open a recent project. As for the implementation of the reconstruction, the process is divided into three main sections: Surface Measurement, Sensor Pose Estimation and Smoothing, and Mesh Reconstruction. Surface Measurement consists of the scanning and obtaining several point clouds from the Kinect. For this scanning process, a turntable was used in order to obtain around 37 point clouds with an approximate interval of 10 degrees per point cloud scan. This data is filtered by a pass through filter. For the Sensor Pose Estimation, the Iterative Closest Point (ICP) algorithm is performed in order to register the obtained point clouds from the Kinect into one. For removing noise and sharp edges from the surface, smoothing is performed by Moving Least Square (MLS) and Laplacian Smoothing. Finally, for reconstructing the 3D mesh, two options out of Greedy Triangulation or Grid Projection is performed. The Poisson reconstruction method was also studied and implemented in this paper, but as the algorithm was the slowest among the ones we used and required an extra cleaning of the 3D point cloud, this method was discarded. This paper includes a detailed explanation of the mentioned techniques as well as some examples of images to demonstrate the final output obtained.

## 2. Choice of Sensor and Related Libraries used

The Kinect is a light laser scanner that generates real time depth maps. It obtains a coloured 3D point cloud, also known as RGB-D image, which provides information about the colour and the depth of each scanned pixel [7]. The Kinect consists of three main elements:

- IR light emitter: beams IR rays light into the scene.
- Monochrome CMOS sensor
- RGB camera: traditional colour camera.

The Kinect converts the pattern obtained by the CMOS sensor into a depth map. The depth map is aligned to the colour picture, obtained by the RGB camera, and then are combined into a single RGB-D image that is sent towards the USB port [8]. Different tools, libraries and platforms were used in this project including Microsoft SDK, OpenNI 2.0, PCL 1.8 and Arduino. In this paper, we will talk briefly about few of them.

### *2.1 Microsoft SDK*

A Software Development Kit (SDK) is a set of tools used to develop software for a determined operating system. Microsoft Windows SDK, is an SDK from Microsoft that provides tools, documentation, header files, libraries, code samples and compilers that developers can use to create applications that run on Microsoft Windows. Windows SDK can be used with the Kinect sensor to develop applications that support gesture and voice recognition.

### *2.2 Open NI 2.0*

The OpenNI 2.0 Application Programming Interface (API) provides access to PrimeSense compatible depth sensors. It allows an application to initialize a sensor and receive depth, RGB, and IR video streams from the device. It provides a single unified interface to sensors and .ONI recording created with depth sensors. OpenNI also provides a uniform interface that third party middleware developers can use to interface with depth sensors. Applications are then able to make use of both the third party middleware, as well as underlying basic depth and video data provided directly by OpenNI.

### *2.3 Point Cloud Library (PCL)*

PCL is an open source project for point cloud processing written in C++. It is aimed to be a cross platform library which is why it can be successfully compiled and deployed on Linux, Windows, MacOS and Android [14]. The PCL framework contains numerous state of the art algorithms for point cloud processing, including filtering, feature estimation, surface reconstruction, registration, model fitting and segmentation. These algorithms can be used, for example, to filter outliers from noisy data, stitch 3D point clouds together, segment relevant parts of a scene, extract key points and compute descriptors to recognize objects based on their geometric appearance, create surfaces from point clouds and visualize them [19]. Since PCL provides all required functionality from the data acquisition to the surface creation and because it is free for commercial and research use, it was the selected library for the implementation of this project.

## 3. Implementation

Implementation of the overall software and the methods used for the development of the whole system is given in this section. Starting from the automation with the turning platform, to the final 3D model, all the methods are sequentially explained in the following subsections. The complete end-to-end flowchart is given in Figure 3.1.1.

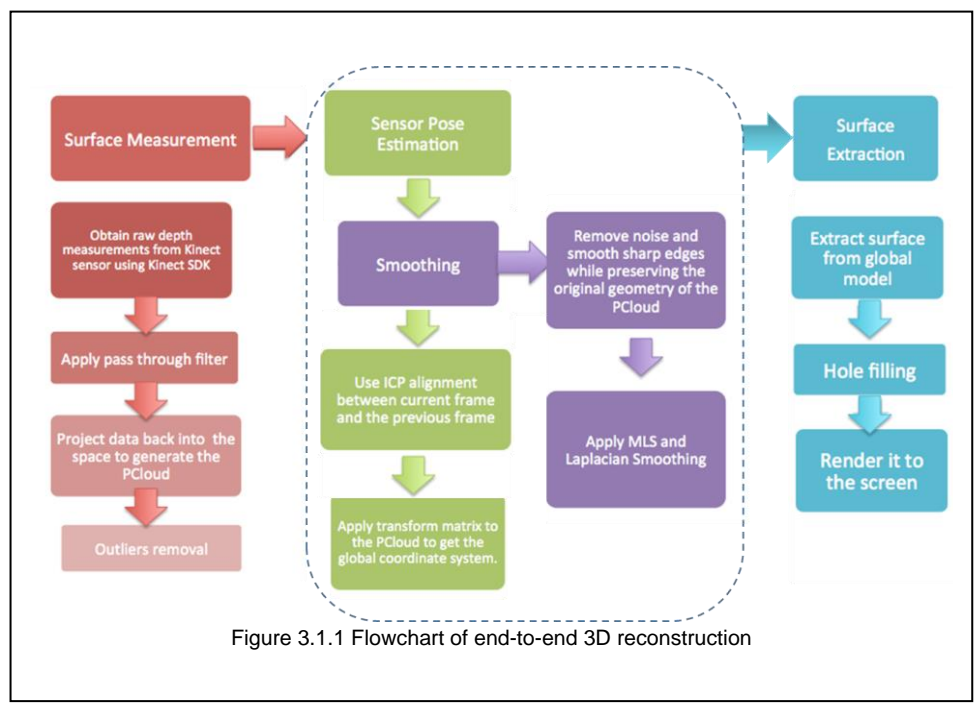

Figure 3.1.1 Flowchart of end-to-end 3D reconstruction

### *3.1 Designing the Turning Platform*

The project included constructing an automated turntable (see Figure 3.1.2) along with our software. An integrated turntable allows a more accurate way of taking point clouds and provides clouds which are already aligned to some extent. This is because the turn table rotates around a pivot and basically this means the object being scanned stays stationary and is rotating around a fixed axis. This also allows us to have a simple click to initiate function and it lets the software and turntable to take care of the rest.

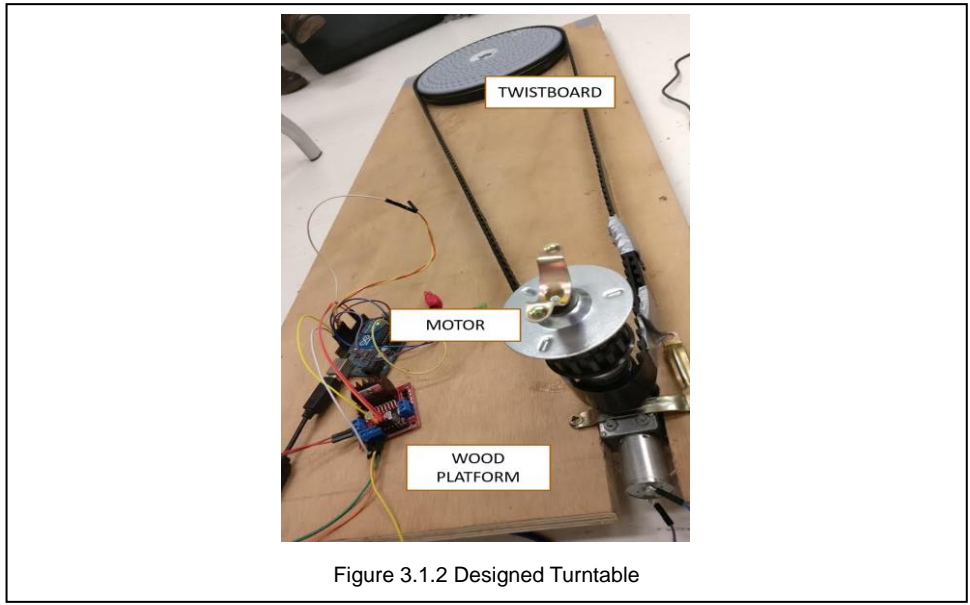
Figure 3.1.2 Designed Turntable

For the mechanical part of building the turntable, the plan was to make a simple turntable that can support and perform flawlessly without much complications. As the main goal of this project is to devise an efficient and cheap method of 3D scanning, the turntable was designed keeping cost in mind. The idea was to use a simple twist board, as the twist board is already designed to support a person. A twist board is basically a round platform which rotates all the way to 360 degrees and it is used mainly for exercising purpose. The existing twist board was modified so a belt can be used to rotate it. For powering this twist board, a 12 V high torque 8 RPM DC gear motor was used. To control the motor, a L298N dual motor controller module was used. An industrial size belt was used to convey power from the motor to the twist board. Since the idea was to introduce an automated system that is user-friendly and can fully communicate with the software, we opted to use Arduino Microcontroller for controlling the motor from the finished software. Serial communication was used to talk back with the Arduino using our software. For controlling the motor, a very simple control function was devised. Additionally, a green LED and a red LED was added to give a physical visual feedback as can be seen from Figure 3.1.3. The serial input takes three values from the software.

1. When our software sends '1' through the serial port, the motor pin output is set to 'high' and also the pin for green LED is set to 'high' and the pin for red LED is set to 'low'. This means both the motor and green LED turns on and red LED turns off when a '1' is passed.

2. When our software sends '0' through the serial port, the motor pin output is set to 'low' and also the pin for red LED is set to 'high' and the pin for green LED is set to 'low'. This means both the motor and green LED turns off and the red LED turns on when a '1' is passed.

3. For a measure of safety, an extra condition was set, that is when our software

sends '2' to the Arduino, all pins are set to 'low'. Meaning everything in the turntable turns off.

4. An additional reset button is included to reset the turntable to its starting position.

The function for capturing point cloud and rotating the turntable is integrated so they can work together hence meaning a full automation of grabbing point cloud data is achieved. In the acquisition part of the software, a 'timer' was used as a measure of control for the increments taken by the turn table. After every 1.8 seconds, '0' was sent to the Arduino to stop the turntable. A point cloud of that frame was taken and then again a '1' was passed to turn on the turntable for 1.8 seconds again.
This way point clouds were taken at every 1.8 second increments until a full 360 degrees of point cloud was obtained. Considering the speed of the motor, for every 1.8 seconds, it was observed that the turntable makes around 10 degrees of increment irrespective of the weight of the object. Hence, the software saves around 36 points clouds. '2' is sent at the end of the acquisition to stop the turntable.

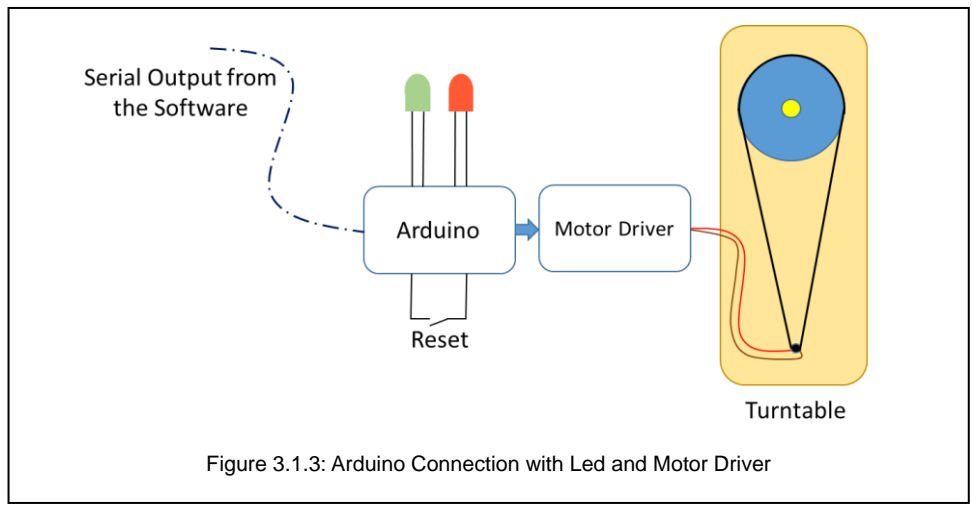

Figure 3.1.3: Arduino Connection with Led and Motor Driver

## 3.2 Overview of Graphical User Interface (GUI)

For the software designed in this paper, a simple GUI was designed for ease of use. The GUI allows the user to take scans, enter parameters, generate 3D models, and also view previously made 3D models. QT does not have a widget dedicated to displaying point clouds. Therefore, Visualization Toolkit (VTK) was integrated with QT to display point clouds and mesh files. The GUI designed has different sets of windows which allow user to visualize and set the pass through filter window using sliders, it allows a user to register the set of point clouds under the models name, and finally it allows to reconstruct the point cloud to a final model using two techniques as preferred by the user. The different windows of the GUI are shown in Figure 3.2.1.

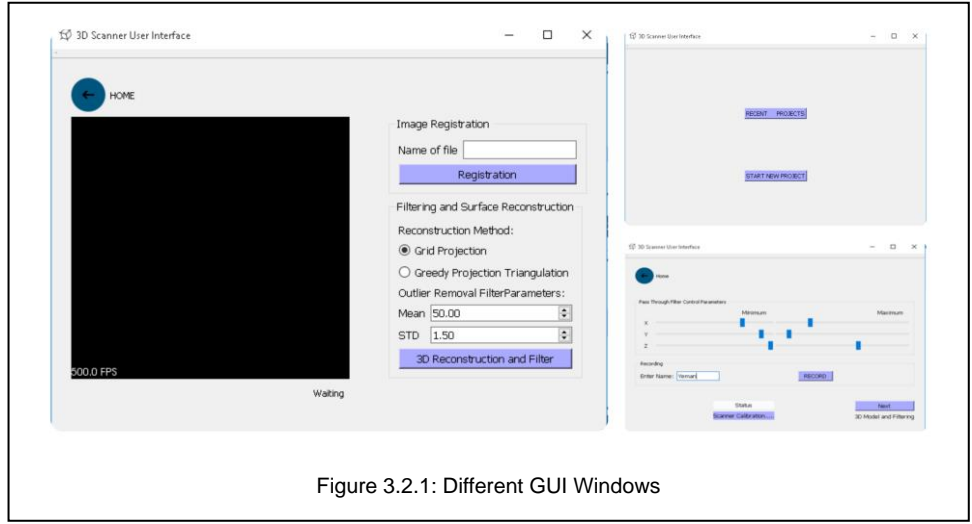

Figure 3.2.1: Different GUI Windows

*3.3 Surface Measurement*

Errors of depth data of the Kinect can make the reconstructed surface jittery if they are not reduced. These errors can be categorized in:

1. Gross error when edges of object surface are scanned: The infrared light emitted from Kinect sensor will not be well reflected when encountering some sharp object or the edge of a surface. This will result in an unreliable data of the edge, e.g. a smooth linear edge appears to be unsmooth in the depth image obtained from Kinect. This kind of gross error will severely harm the result because one cannot assure whether the points near the edge is behind or in front of the surface.

2. Gross error when objects that are not well reflective are scanned: When encountering some specular objects such as metal or glass (e.g. the monitor of a computer), Kinect sensor can hardly detect the depth data of these surfaces because these types of material do not reflect the infrared light, so no structured-light depth reading is possible. As a result, the depth image contains numerous holes where the data are missing.

3. Small Gaussian noise: Obtained from the depth map. A pre-process to reduce the noises and eliminate the gross error is required every time a raw depth map is obtained. This pre-process consists of applying a Bilateral Filter, which reduces the noise and closes the small holes while preserving the details of the image.

To remove as much noise before the acquisition, a pass through filtering technique is used before capturing the point clouds.

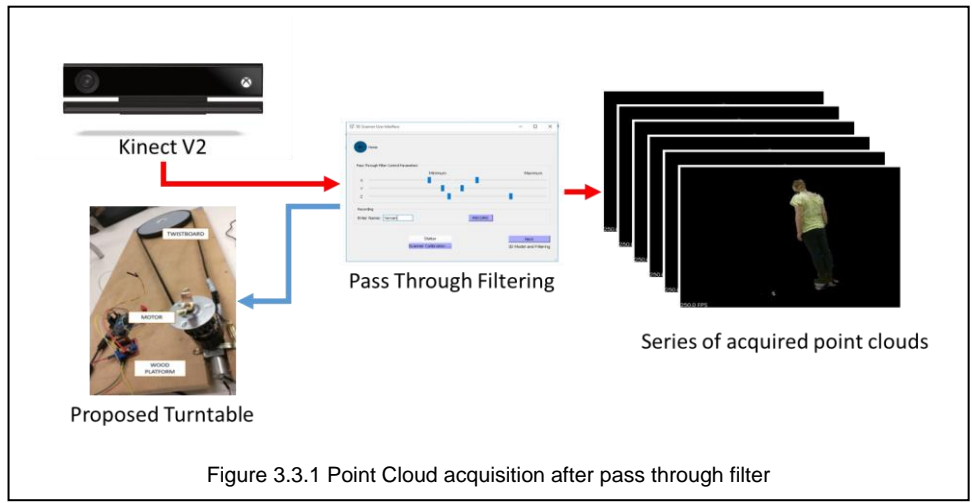

Figure 3.3.1 Point Cloud acquisition after pass through filter

This PassThrough Filter (PTF) removes the parts of the point cloud which lie outside the defined range in the three X, Y and Z direction. In this project, the filter is applied while acquiring raw data (in real time), moreover, the user is given the choice to define the range which they want to keep. From the GUI, the user can select a virtual bounding box by moving the sliders, where, only the data inside the bounding box is taken and the data outside the bounding box is omitted. The filtering window drawn can be visualized clearly in Figure 3.3.2.

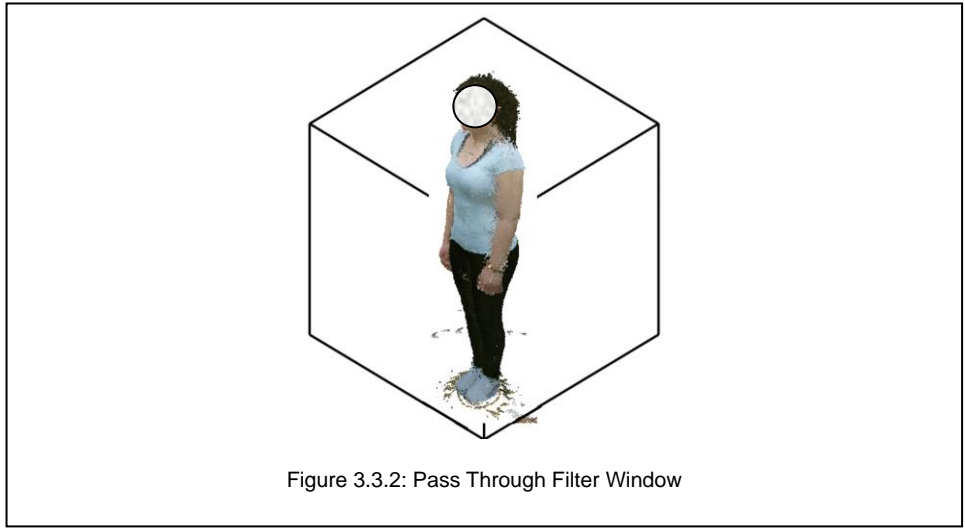

Figure 3.3.2: Pass Through Filter Window

*3.4 Sensor Pose Estimation and Smoothing*

The second main step for achieving our 3D reconstruction is the sensor pose estimation. Pose estimation consists of finding the transformation of an object in a 2D image so that the 3D object can be reconstructed from it. After finishing scanning with the Kinect, a series of point clouds are generated. This point clouds need to be registered and merged into a 3D model. To achieve this, estimation of the 3D rigid transformation matrix of the sensor pose is needed and transformation of the point cloud into the global system coordinate is also needed [9]. In this work, the estimation of the sensor pose was done using the ICP algorithm.

*3.4.1 Overview of Iterative Closest Point (ICP)*
The ICP algorithm is used for geometric alignment of 3D models when an initial estimate of the relative pose is known [10]. In the ICP algorithm, one point cloud, the reference, is kept fixed, while the other one, the source, is transformed to best match the reference. The algorithm iteratively revises the transformation needed to minimize the distance from the source to the reference point cloud. After scanning a person, the next task is to combine the consecutive scans into a single point cloud. To achieve that, the first frame is fixed as a reference of the coordinate system to be used. All the other frames will be transformed into that coordinate system. This transformation can be calculated iteratively using the ICP algorithm which will give, as a result, the alignment of the successive individual scans of point clouds into a unified 3D global map [11].

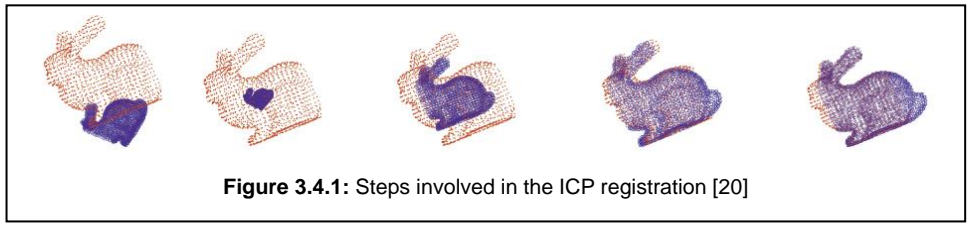

**Figure 3.4.1:** Steps involved in the ICP registration [20]

The basic principle of ICP algorithm is to find the rotation and translation parameters between two point clouds during iterations [7]. The general steps of the algorithm are the following. First, a reference point cloud must be chosen $P_0$. Second, given a sequence of point clouds $P_i$, the ICP algorithm finds the transformations $T_i$, such that

$$P_{i-1} \approx T_i P_i$$
*where, i = 1,..,N*

Applying transformations iteratively, the cumulative transformations $TC_i$ can be found by:

$$TC_{i-1} = T_i * TC_{i-1}$$
*where, i = 1,..,N*

Cumulative transformation matrices can be used to bring all point clouds into the same coordinate reference:

$$Q_i = TC_i * P_i$$
$$\text{where, } i = 1,..,N$$

Finally, the reconstructed point cloud Q is formed by the union of all aligned point clouds.

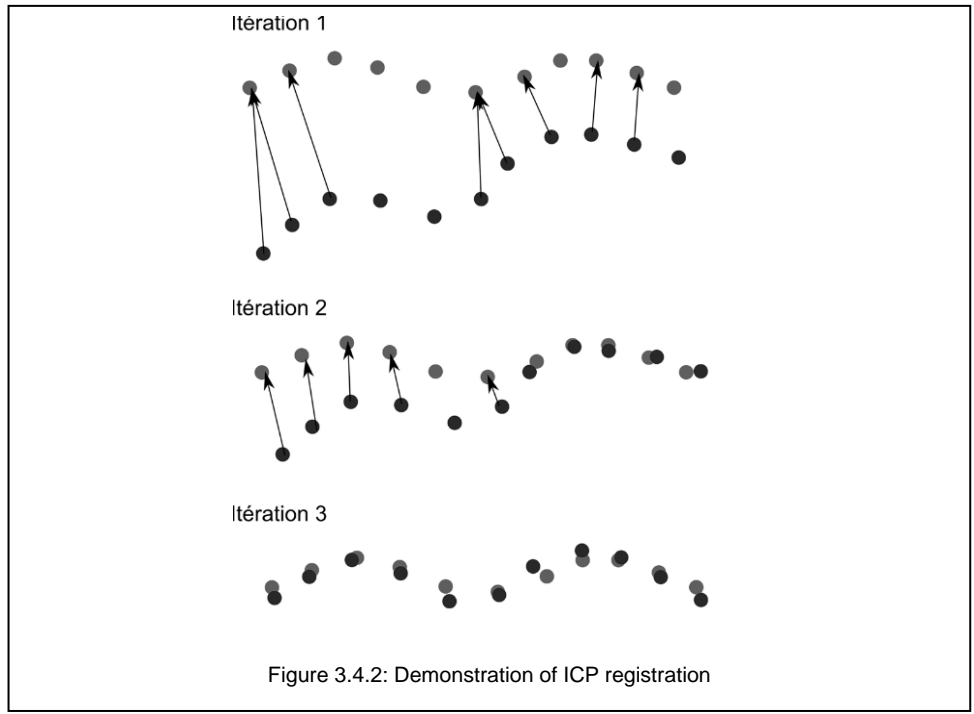

Figure 3.4.2: Demonstration of ICP registration

### 3.4.2 Statistical Outlier Removal Filter (SOR)

Measurement errors cause sparsing outliers which destroy the results of scanning. This makes the estimation of the PCL characteristics like surface normal. It is also leading to erroneous values which might cause registration failures. These issues can be managed by applying a statistical analysis on each point's neighbourhood and removing those which do not meet certain conditions. The SOR (Figure 3.4.3) we are applying is based on the computation of the distribution of point to neighbour's distances in the input dataset. For each single point, it computes the mean distance from it to all its neighbours. As the resulted distribution is Gaussian with a mean and a standard deviation- all points whose mean distances are outside an interval defined by the global distances mean and standard deviation can be considered as outliers and trimmed from the dataset.

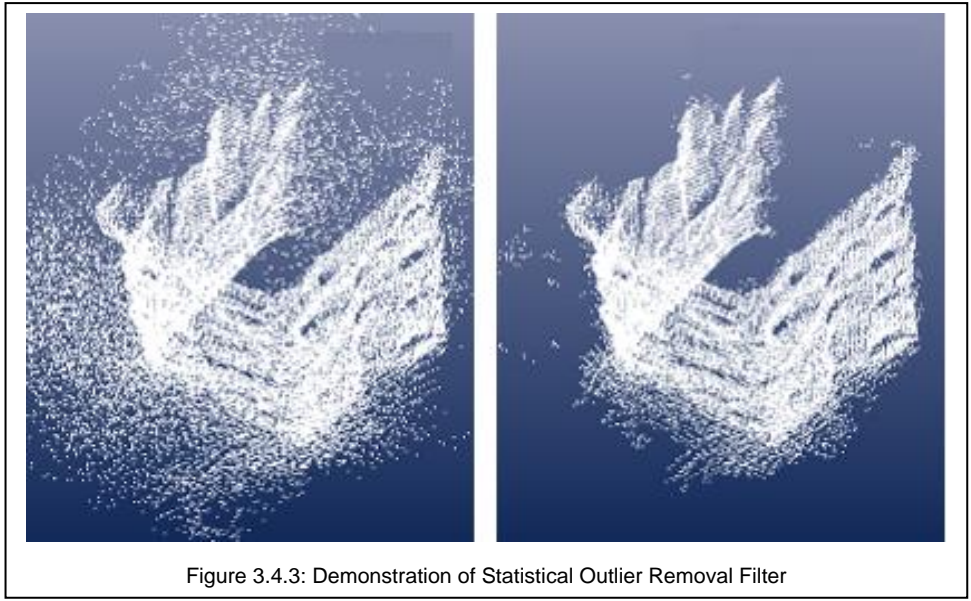
Figure 3.4.3: Demonstration of Statistical Outlier Removal Filter

The number of neighbours to analyze for each point is set to 50, and the standard deviation to 1. This was observed to give the best result as seen from the research conducted. This means that all points that have a distance larger than one standard deviation of the mean distance to the query point will be marked as outliers and removed.

### 3.4.3 Moving Least Square Smoothing (MLS)

The second step after filtering is smoothing the point cloud using MLS method. The MLS is used to smooth and resample the noisy data. The filters applied before were not completely effective in removing the irregularities in our point cloud as some of these irregularities are very hard to be removed using SOR. In addition to remove the irregularities, MLS can help in creating complete models by resampling the data which recreate the missing parts of the surface in our point cloud by using higher order polynomial interpolations between the surrounding data points. As a conclusion, resampling using 'double walls' can fill the missing parts in our point cloud and then the artefacts caused by registering multiple scans together, so the resultant point cloud will be a smoothed one. From Figure 3.4.4, the effect can be visualized clearly, the image on the left has alignment errors hence the surface normals are noisy. After applying MLS, the image on the right is observed, where, the surface normal can be seen to be more homogenous.

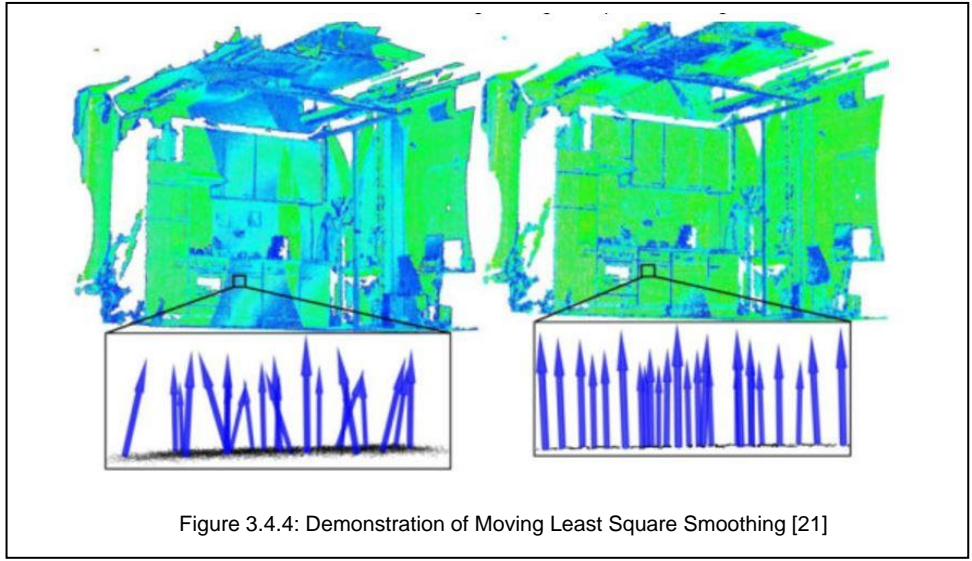

Figure 3.4.4: Demonstration of Moving Least Square Smoothing [21]

### 3.4.4 Laplacian Smoothing (LS)

In our project we apply two smoothing methods, the first one is MLS which we mentioned above to smooth and resample the 3D point cloud. The second smoothing method is the LS which we apply on the mesh created by greedy triangulation reconstruction. In LS, first, all the vertices are being searched and then all the vertices are connected to the searched ones are being found as well. These informations are stored in an array which is being updated with time. For the corner vertices, normally no smoothing is applied, but for the edge vertices, they are smoothed only along the edge, in addition, they are smoothed if the angle of the edge is smaller than a specific threshold. For all the other vertices, their new locations are being set by taking the average of the location of the other vertices connected to them.

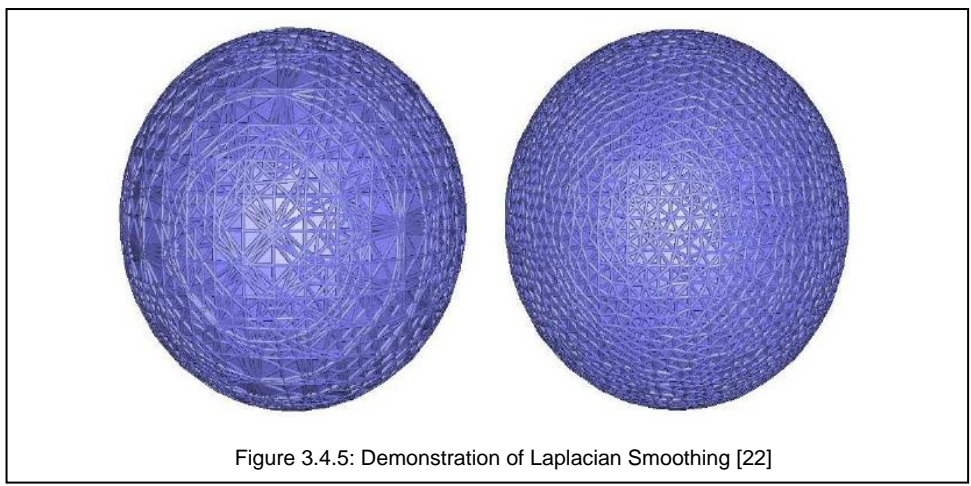

Figure 3.4.5: Demonstration of Laplacian Smoothing [22]

After storing the point clouds of the object taken from different angles individually in the file directory, they are called back and vectorized and they are fed as an input to the alignment function. The alignment process follows the following steps:

- The first point cloud at zero degrees is taken as the source and the immediate second point cloud as the Target for alignment.

- The point clouds are down sampled to speed up the computation and also to over complicating the sensor pose estimation. A threshold is set at the leaf size which regulates the amount of the down sampling done to the point cloud without losing critical information of the geometry of the object itself.

- Before feeding in the two down sampled point clouds for alignment, outliers and noise are removed using the statistical outlier removal function. A tolerance limit is set against the mean and if any point lies outside the limit, the point is considered as an outlier, otherwise it remains.

- Once the outliers and noise are removed, the ICP algorithm is performed on the Source and Target point cloud, and the two clouds after transformation are concatenated and then passed as a source for the next iteration.

- This way all the point clouds from the set of acquired point clouds from the Kinect are aligned and combined to form a final point cloud for the 3D model.

Before the final point cloud is passed over to the next step for Mesh Reconstruction, a series of MLS and LS is performed to further smooth the final point cloud. The parameters of the smoothing functions can be set by the user or the user can use the default values provided by the software.

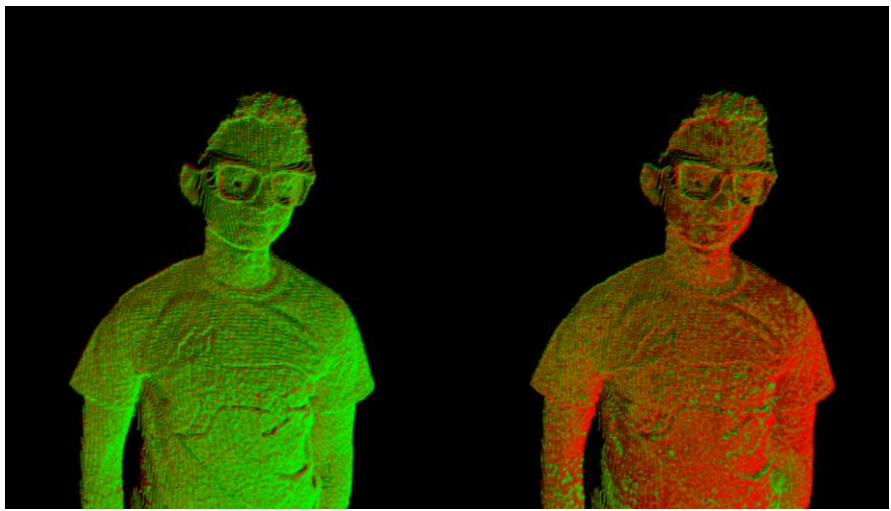

Figure 3.4.6: Demonstration of ICP applied on aligning point clouds

*3.5 Mesh Reconstruction*

After the output from the sensor pose estimation, an aligned point cloud of the object being scanned is acquired. Now a series of algorithms are performed in order to convert the input point cloud data into a constructed polygon mesh. Three algorithms were implemented in this work:

1. Greedy Triangulation
2. Grid Projection
3. Poisson

*3.5.1 Greedy Triangulation (GT)*
GT is an algorithm that receives a point of clouds and converts it into a triangle mesh by projecting the local neighbourhoods of a point along the point's normal and connecting unconnected points. We perform GT to create a triangular mesh from the point of clouds obtained from the sensor pose estimation part. The focus of GT is to keep a list of possible points that can be connected to create a mesh. GT is obtained by adding short compatible edges between points of the cloud, where these edges cannot cross previously formed edges between other points. GT can, however, over-represent a mesh with triangles when these are represented by planar points. GT contains accurate and computationally efficient triangulations when the points are planar. However, a big difference can be noticed when the triangulations are non-planar, this can be fixed by increasing the number of boundary vertices or linearly interpolating between the boundary vertices [12].

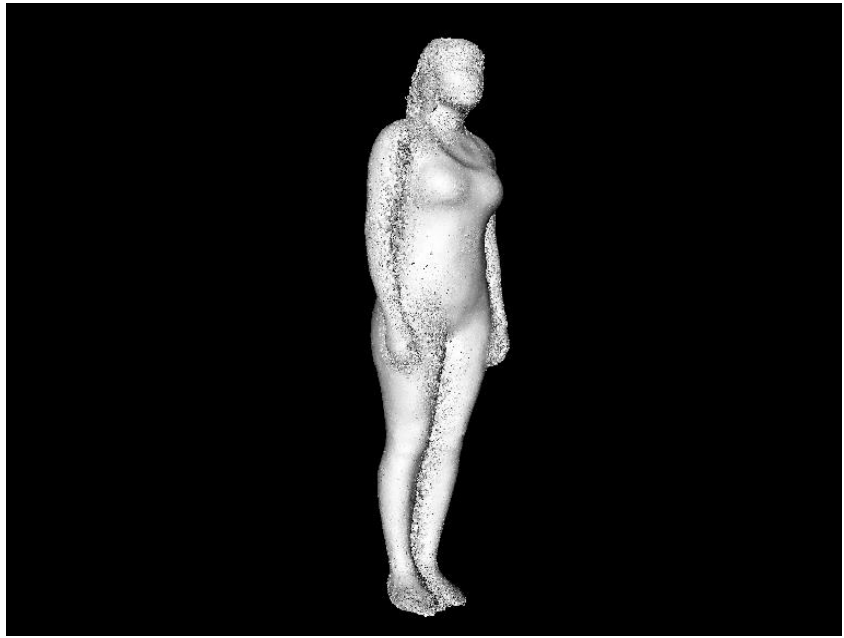

Figure 3.5.1: Model using Greedy Triangulation

An approach at computing the GT is to compute all distances, sort them and examine each pair of points in length and compatibility with edges created. The Greedy Triangulation algorithm is simple at its core but also not very reliable. With a point cloud, S, of n points, the algorithm looks for the closest point where a compatible edge can be created between the two. A compatible edge can be described as an edge between two points that does not intersect with any other edge [13]. In the Point Cloud Library, Greedy Projection triangulation works locally, however, it allows the specification of the required features to be focused on. Such parameters are neighbourhood size of the search, the search radius and the angle between surfaces [14]. As seen from Figure 3.5.1, Greedy Triangulation Model provides a more detailed but edgier output. This can be further reduced using more filters and smoothing.

### 3.5.2 Grid Projection (GP)

The second mesh reconstruction algorithm implemented in this work was GP. GP tries to create seamless surfaces that lack boundaries or proper orientation. One of the defining points for GP is that a continuous surface is created, as opposed to Greedy Triangulation, where the surface is not continuous and can contain holes [15]. We perform Greedy Projection to create a rectangular mesh, without holes, from the smoothed point of clouds. The GP algorithm is a grid based surface reconstruction algorithm. Points are first partitioned into voxels, and a vector field is constructed, where the vectors at any given point are directed at the nearest point. A surface is then determined by examining where vectors with opposite directions point towards. Edges in the voxels that this surface are reconstructed from are determined and padding cells (cells neighboring the voxels containing the points) are also added. The center points of each voxels are then projected based on the edge intersections, and the surface is reconstructed by connecting these center points [16].

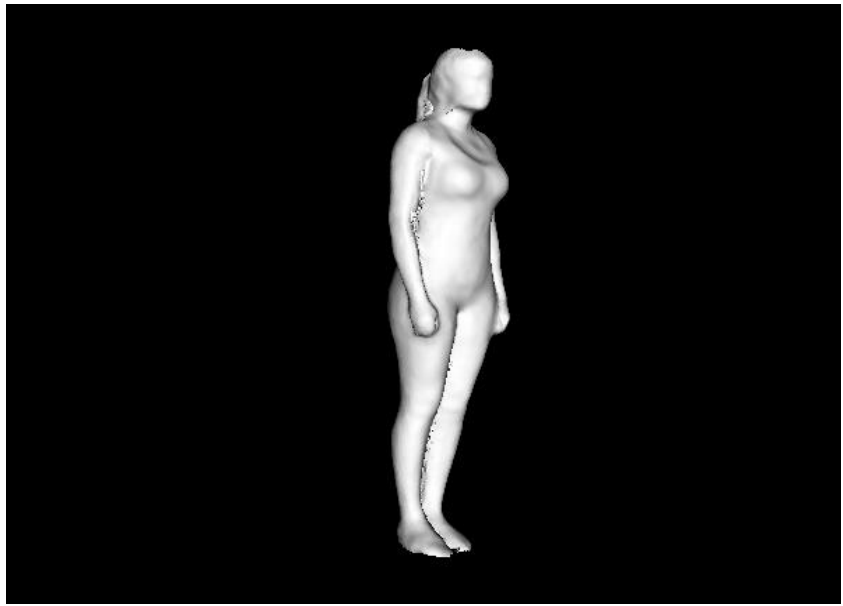

Figure 3.5.2: Model using Grid Projection

The method requires only two parameters: (1) padding and (2) resolution. The output of GP of our program can be seen in Figure 3.5.2. As seen from Figure 3.5.2, Grid Projection Model provides a smoother and better-looking model.

*3.5.3 Poisson*

Poisson is a mesh reconstruction from oriented points which can be cast as a spatial Poisson problem [17]. We perform the Poisson algorithm to obtain a watertight mesh from a set of smooth and filtered point of clouds. The Poisson reconstruction algorithm reconstructs a watertight model by using the indicator function and extracting the iso-surface. The indicator function is a constant function, obtained from the sampled data points, whose computation would result in a vector with unbounded values. The surface integral of the normal field is approximated using a summation of the point samples and then reconstructed using its gradient field as a Poisson problem [18]. The gradient field is used to reconstruct the indicator function of the surface. The Poisson reconstruction uses the indicator function that works best with noisy data points and can, therefore, recover finer details [18]. The biggest problem with Poisson and the reason why it is not usable for this project is the fact that it requires a clean mesh, without noise to actually reconstruct the object of interest. Therefore, in order to use Poisson, a lot of manual clean up should be done before trying to actually apply the algorithm. Figure 3.5.3 includes an example of Poisson reconstruction using our data with a depth of 10.

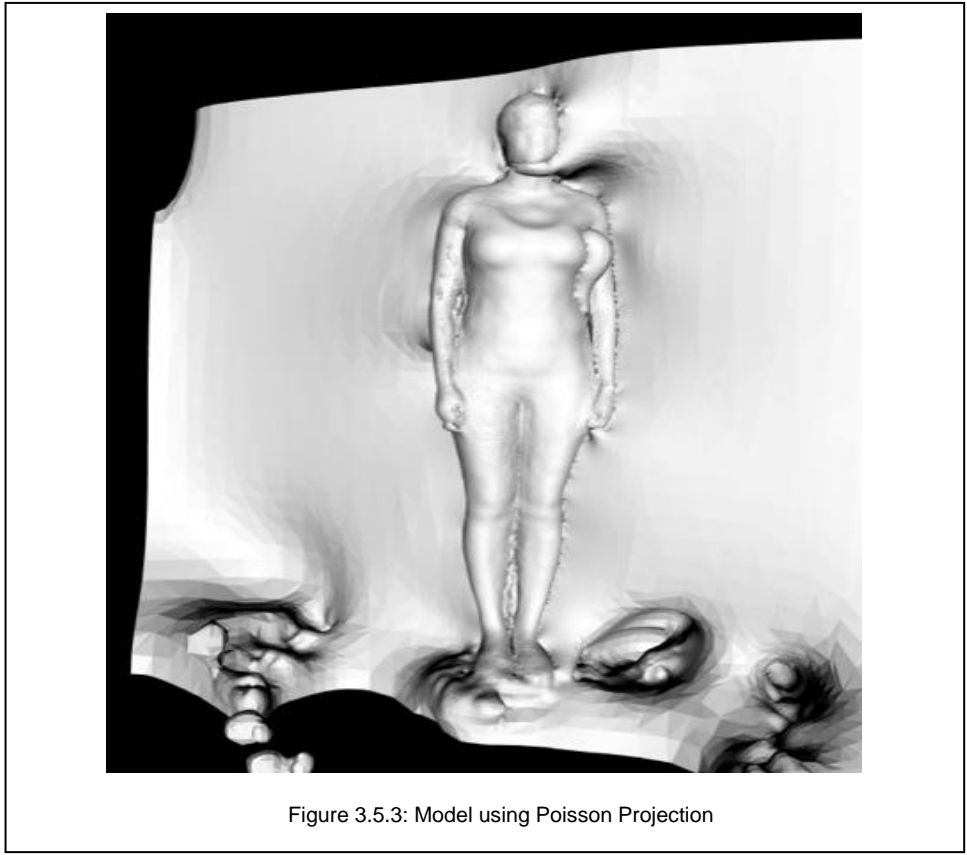

Figure 3.5.3: Model using Poisson Projection

*3.5.4 Hole Filling*

Even after using the best combination of techniques, the final model still produces some holes in it. The origins of these holes ranged from early scanning issues, where some parts of the body can't be scanned properly (lie the bottom of our feet), to later stages, where filtering and meshing algorithms can distort the surface of the model and create additional holes. In case of usage of Grid Projection, amount of holes drastically decreases, but at the same time, the orientation of the squares, which are created by the algorithm (i.e. angles and orientations) can create small holes between a tiny number of squares, ranging from one. This can be fixed by usage of the built-in VTK library – vtkFillHolesFilter, which can fill holes with following types of edges:

1. Boundary (used by one polygon) or a line cell;
2. Non-manifold (used by three or more polygons);
3. Feature edges (dihedral angle > than a threshold);
4. Manifold edges (edges used by exactly two polygons).

But, nevertheless, this library is rendered problematic for big holes. Most of the time, hole issue occur nearby and inside the seams and joints, which can be clearly seen in the Figure 3.5.4.

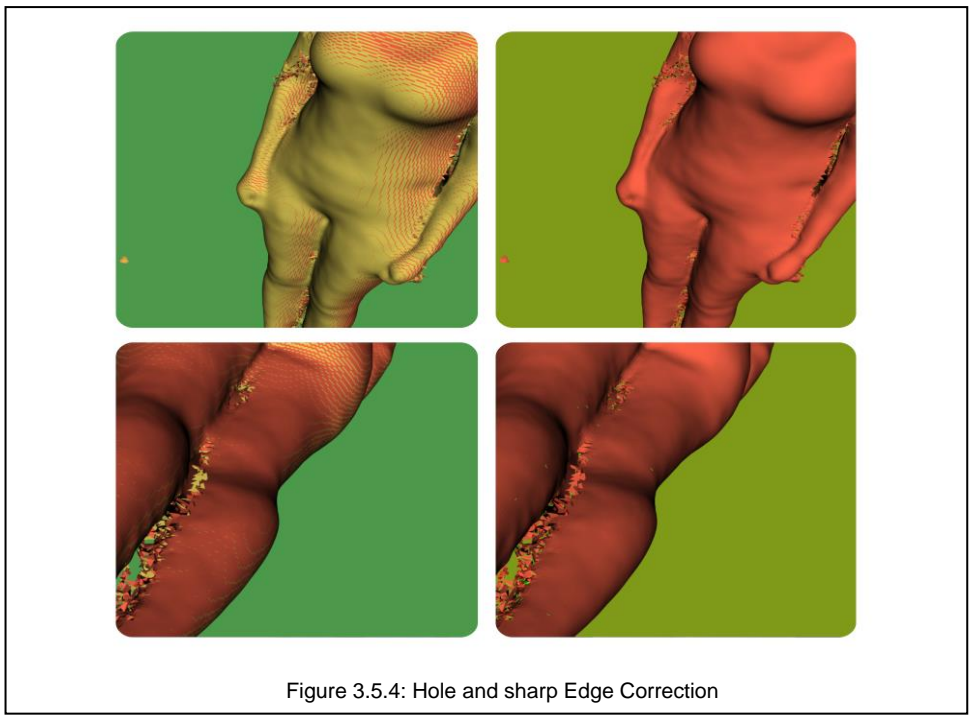

Figure 3.5.4: Hole and sharp Edge Correction

Therefore, apart from the hole detection which can be seen on the pictures, we propose a pipeline approach, which sums up to extraction of the detected edges and filling them with created patches, which can be passed through vtkTriangleFilter in case of usage of the Grid Projection method. This pipeline approach potentially solves the problems of the vtkFillHolesFilter, closing the big holes on the feet and head. It can be a suitable direction for the development of our project.

Pipeline approach comprises four steps:

1. Extraction of the edges (vtkFeatureEdges, vtkExtractEdges)
2. Create patches (vtkStripper)
3. Use contours as an input and fill holes with created patches (vtkTriangleFilter)
4. Append changes to the mode

## 4. Results

To perform final scans from the Kinect, the turntable was set to turn every 1.8 seconds, with an increment of approximately 10 degrees. After a full 360 degree rotation is completed, the turntable stops and the Kinect stops taking scans. In total, around 36 point clouds are generated from this process. An important remark to be made is that, although the software achieved its purpose of scanning and performing 3D reconstruction of a person, the final 3D model was lacking some fine details. This was due to the fact that during the scanning, the Kinect was positioned at least one meter away from the person. By increasing the distance between the Kinect and the person, a lot of information was lost and more noise is introduced. When the time to perform the reconstruction came, only a low percentage of data was left to work with. To remove noise and unwanted data a real time pass through filter is applied in order to delimit the region of scanning, ignoring the background and focusing only on the person to be scanned. After that, an outliers' removal is performed with a standard deviation value (STD) of 0.5. To merge all the scans into one single point cloud, the ICP algorithm is performed. Both before and after obtaining the reconstructed 3D point cloud, the MLS algorithm is applied in order to make the surface smoother. Once the surface is smoothed, the mesh is extracted. We applied two different methods: GT and GP. Figure 4.1 shows the result of applying GT with mu = 5, D = 50 and R = 0.025.

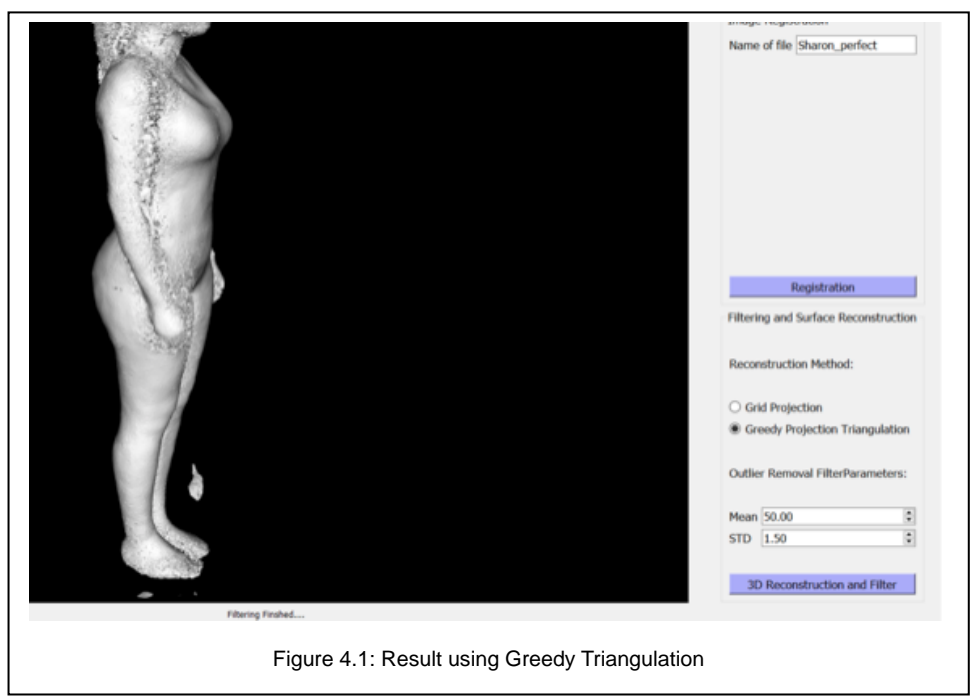

Figure 4.1: Result using Greedy Triangulation

The second reconstruction technique was GP. It is a method that requires more computational time than GT but gives as output a rectangular mesh without holes, requiring no smoothing afterwards. Figure 4.2 shows the results of GP with a resolution (res) of 0.0025.

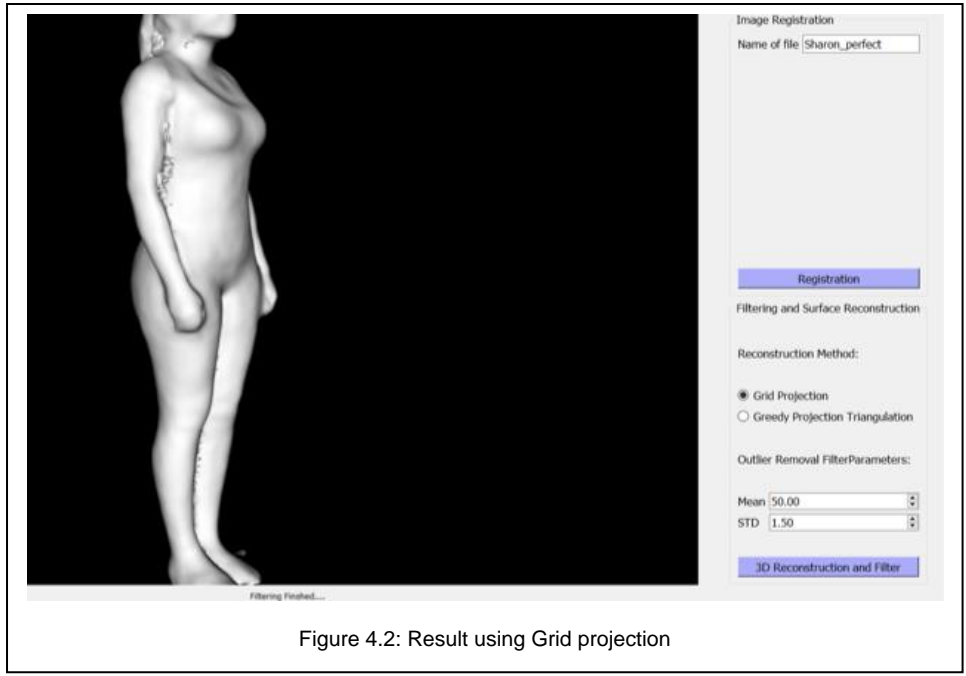

Figure 4.2: Result using Grid projection

Another method we tried to implement was Poisson. It is a method that gives a watertight mesh with fine details but is slower than GP and GT and also amplifies the noise, requiring to do a lot of previous cleanup of the mesh before trying to implement it. This is the reason why the Poisson method was discarded for the final implementation of our software.

## 5. Conclusion

The designed software can successfully scan a person using a low-cost sensor, reconstruct a 3D model from the scans and export it for 3D printing. The source code for the fully developed software can be found in our GitHub repository, www.github.com/tazleef/Software-Engineering-Project.

During the implementation of this project, a thorough research was performed in order to narrow down and find the final materials and methodologies which gave us the most optimal result. First of all, we had to decide which library to utilize. For this project, the PCL was chosen as the best option for achieving our goals. Second, as multiple research has been performed with Kinect and PCL, the Kinect v2.0 was selected as the ideal sensor for the project. Then, for cleaning and filtering the data, a real time pass through filter and a statistical outlier removal was applied. For combining the consecutive scans into a single point cloud, the ICP algorithm was performed. For obtaining a better shaped and evenly distributed mesh, MLS and Laplacian were used as smoothing techniques. Finally, for reconstructing the 3D model two algorithms were implemented: Grid Projection and Greedy Triangulation. An improvement that can be made for this project, is to perform the scan by positioning the Kinect closer and building a pulley rig for the Kinect to go up and down while scanning.

## ACKNOWLEDGMENT

This work was a semester project for the software engineering course at University of Burgundy, France. Authors wish to thank Katherine Sheran, Hassan Saeed, Islam Mokhtari, Ivan Mikhailov and Solene Guillaume for their contributions to this project including coding and gathering information for the final software. In addition, authors would like to thank warmly Prof. Fougerolle Yohan and PhD. Cansen Jiang for their supervision and guidance. This work was supported by a grant from Mr Alexander Hermanns which made purchasing the camera sensors and the turntable equipment possible for us.